# BigGAN-based Bayesian reconstruction of natural images from human brain activity

Kai Qiao,[1] Jian Chen,[1] Linyuan Wang,[1] Chi Zhang,[1] Li Tong,[1] Bin Yan[1]

[1] PLA strategy support force information engineering university, Zhengzhou, 450001, China.

Correspondence should be addressed to Bin Yan; ybspace@hotmail.com

## Abstract

In the visual decoding domain, visually reconstructing presented images given the corresponding human brain activity monitored by functional magnetic resonance imaging (fMRI) is difficult, especially when reconstructing viewed natural images. Visual reconstruction is a conditional image generation on fMRI data and thus generative adversarial network (GAN) for natural image generation is recently introduced for this task. Although GAN-based methods have greatly improved, the fidelity and naturalness of reconstruction are still unsatisfactory due to the small number of fMRI data samples and the instability of GAN training. In this study, we proposed a new GAN-based Bayesian visual reconstruction method (GAN-BVRM) that includes a classifier to decode categories from fMRI data, a pre-trained conditional generator to generate natural images of specified categories, and a set of encoding models and evaluator to evaluate generated images. GAN-BVRM employs the pre-trained generator of the prevailing BigGAN to generate masses of natural images, and selects the images that best matches with the corresponding brain activity through the encoding models as the reconstruction of the image stimuli. In this process, the semantic and detailed contents of reconstruction are controlled by decoded categories and encoding models, respectively. GAN-BVRM used the Bayesian manner to avoid contradiction between naturalness and fidelity from current GAN-based methods and thus can improve the advantages of GAN. Experimental results revealed that GAN-BVRM improves the fidelity and naturalness, that is, the reconstruction is natural and similar to the presented image stimuli.

## Keywords

## Introduction

Human brain decoding (Cox and Savoy 2003; Haxby, et al. 2001; Haynes and Rees 2006; Kamitani and Tong 2005; Mitchell, et al. 2004) is an interesting topic in neuroscience. Functional magnetic resonance imaging (fMRI) (Ogawa, et al. 1990) can effectively reflect brain activity and human vision (Logothetis and Sheinberg 1996; Ungerleider 1982) is the major sensory channel to acquire external information. Thus, human visual decoding through fMRI data has recently received increasing attention (Horikawa, et al. 2013; Huth, et al. 2012; Naselaris, et al. 2011; Norman, et al. 2006). In contrast with visual encoding (Kay, et al. 2008; Mitchell, et al. 2008; Wen, et al. 2018; Zhang, et al. 2019) that predicts brain activity in response to visual stimuli, visual decoding aims to predict the content of image stimuli through brain activity. According to different targets, visual decoding can be divided into classification, identification, and reconstruction (Naselaris, et al. 2011). Reconstruction (Hossein-Zadeh 2016; Lin, et al. 2019) of image stimuli (usually called visual reconstruction) requires full-information decoding and is the most difficult task. High visual reconstruction quality which indicates the degree of human vision understanding, has been studied.

Previous methods mainly focused on the reconstruction of simple and small sized image stimuli, such as domino patterns (Thirion, et al. 2006), hand-written digits (Hossein-Zadeh 2016), and English letters (Schoenmakers, et al. 2013). These methods usually considered visual reconstruction as a simple predicting problem because the dimension of predicting space is not high and directly (Miyawaki, et al. 2008) employed linear and nonlinear mapping to accomplish reconstruction. Some methods (Qiao, et al. 2018; van Gerven, et al. 2010) depended on some middle features to construct reconstruction models to further improve the quality of reconstruction. In addition to simple images, Naselaris et al. (Naselaris, et al. 2009) first implemented the reconstruction of natural images by using some priori information and combination of structural (Kay, et al. 2008) and semantic encoding models. The method can be essentially regarded as an image retrieval problem in a limited natural image library and belonged to a very preliminary attempt.

After deep networks, especially convolutional neural network (CNN), are introduced into neuroscience (Güçlü and van Gerven 2015; Yamins, et al. 2014) domain, CNN features have shown a strong correlation with the brain activity in visual cortices (Eickenberg, et al. 2016; Horikawa and Kamitani 2017), and CNN features have been applied to reconstruct natural images from fMRI data. This application has wide used the two-step manner of visual reconstruction. Depending on feature representation, fMRI data is first mapped into the specific layer of CNN features, and second, reconstruction can be achieved by inverting the predicted features into images. For example, Wen et al. (Wen, et al. 2017) mapped fMRI data to the first layer of CNN features and employed deconvolution neural network (Zeiler, et al. 2011) to implement the reconstruction of dynamic video frame by frame. Similarly, Zhang et al. (Zhang, et al. 2018a) mapped fMRI data to the early layer of CNN features, and inverted the features into image space by using feature visualization (Mahendran and Vedaldi 2015) for reconstruction. Han et al. (Han, et al. 2017) mapped fMRI data to the latent feature space of pre-trained variational autoencoder (VAE) (Schmidhuber 2015) and used the decoder in the VAE for reconstruction. These methods successfully reconstructed visual stimuli from fMRI data; however, the reconstruction results were fuzzy, unnatural, and cannot form semantic understanding.

Generative adversarial network (GAN) proposed by Goodfellow et al. (Goodfellow, et al. 2014) is a powerful neural network for image generation and has attracted increasing attention (Arjovsky, et al. 2017; Berthelot, et al. 2017; Chen, et al. 2016; Zhao, et al. 2016). GAN contains a generator and a discriminator. The generator uses random noise vectors sampled from a fixed distribution to generate fake images and tries to fool the discriminator. The discriminator tries to distinguish between real and fake images. The ability of generator and discriminator continually improves by competing with each other during training, and the generator can finally generate images that look like real images. In this way, GAN accomplishes the transferring from random noise distribution to natural image distribution. Various GANs such as condition GAN (CGAN) (Mirza and Osindero 2014) that requires the category prior, and GAN-based applications such as style transferring (Zhu, et al. 2017) have vastly emerged. GAN is currently the best model for image generation. Visual reconstruction is a problem in conditional image generation on fMRI data, and the use of GAN to reconstruct natural images from brain activity has attracted increasing attention.

Current GAN-based visual reconstruction methods are mainly divided into three types. The first type of method (Shen, et al. 2018) replaced the original random noise vector with fMRI voxels as the input of generator and additionally added a fidelity loss item on the original GAN loss to restrict the generated images similar to the corresponding image stimuli. During training, generated images are required natural by the discriminator, and similar to the corresponding image stimuli by fidelity item. However, the distribution transferring may be impaired regarding the unnatural reconstruction results. This result is because the additional fidelity item expects the one-to-one mapping from the fMRI voxels to the corresponding image stimulus instead of the distribution-to-distribution mapping from the fixed noise distribution to natural image distribution. In addition, GAN is difficult to train (Arjovsky, et al. 2017) with the minority of fMRI data samples. The reconstructed image from EEG brain signal based on this manner (Jiao, et al. 2019; Palazzo, et al. 2017; Tirupattur, et al. 2018) also exhibited similar characteristics, and the naturalness and fidelity of reconstruction cannot be simultaneously obtained. For example, EEG-GAN method (Palazzo, et al. 2017) generated some unnatural images of the specified category. The second type of method (St-Yves and Naselaris 2018) regarded fMRI voxels as the condition to feed into CGAN. Similarly, fidelity and natural items were both included in the model loss. Through adversarial training, given one random noise vector and fMRI voxels (condition vector), the generator of CGAN can produce the corresponding image stimulus for reconstruction. The first two types of methods both have difficult in balancing the fidelity and natural items by using the minority fMRI data samples, and the reconstruction results are either unnatural or inconsistent. Hence, the third type of method (Seeliger, et al. 2018) employed the generator of pre-trained GAN, and constructed one linear regression mapping from the fMRI activity space to input noise space. The fidelity item between the generated images and true image stimuli was used to provide the gradient information through pre-trained generator network to update the weights of the linear regression model during training. After training, given the fMRI voxels of one image stimulus, the linear model mapped them to noise space, and the generator can reconstruct an image. This type of method improve the reconstruction results and is based on an assumption that fMRI activity space and fixed noise distribution space have a linear relationship. The inappropriate assumption influences the fidelity of reconstruction, although the reconstruction preserves the naturalness of generated images because it did not need to retrain the generator. Hence, employing the generator of pre-trained GAN instead of retraining GAN may be appropriate for current fMRI data. In conclusion, these GAN-based methods have improved the reconstruction of natural images from fMRI data. However, further improving the fidelity and naturalness of reconstruction is still expected.

Although GAN have obtained enormous progress in image generation, generating diverse normal-sized natural images, such as the famous ImageNet dataset with 1000 categories, are still difficult. Lately, BigGAN (Brock, et al. 2018) can generate amazing natural ImageNet images. In this study, we introduced the best model into the visual reconstruction. Considering that a reconstruction can be defined as the image that has the highest posterior probability of evoking the measured brain activity (Naselaris, et al. 2009), we proposed the new GAN-based Bayesian visual reconstruction method (GAN-BVRM) on one prebuilt visual encoding model and the generator of pre-trained BigGAN to improve the fidelity and naturalness of reconstruction. Essentially, we consider the generator as the infinite natural image dataset and determine the optimal images to fit the encoding model as reconstruction. On the one hand, we can avoid the mutual interference of fidelity and natural item during retraining the GAN by using the pre-trained generator. On the other hand, we did not change the original input noise space and used the encoding model to improve the fidelity. Given that BigGAN model essentially belongs to CGAN and category prior is required for generating images, GAN-BVRM is composed of the four following parts: encoding part that maps image stimuli to brain activity, category decoding part that predicts categories from brain activity, natural image generator of pre-trained BigGAN, and evaluator ranking generated images. By continuous evaluation and search, the improved natural images that maximize to fit the encoding model can be determined as the reconstruction.

In this study, our main contributions are as follows: 1) we analyzed the current drawbacks of visual reconstruction methods based on GAN; 2) we proposed the GAN-BVRM to further improve the fidelity and naturalness of visual reconstruction.

## Materials and Methods

### Experimental data

The dataset employed in our work was based from the previous studies (Kay, et al. 2008; Naselaris, et al. 2009). The dataset had visual stimuli and the corresponding fMRI data that consist of 1750 training samples and 120 testing samples. Voxels in the five regions of interest (ROI: V1, V2, V3, V4, and LO) from low-level to high-level visual cortices were collected in the dataset. In addition to image stimuli and corresponding fMRI data, five experienced persons manually labelled the 1870 images and the final labels with 10 categories (many humans, few humans, mammal, non-mammal, non-building, building, plant, non-plant, organic texture, and inorganic texture) were obtained through voting. The detailed information about the visual stimuli and fMRI data is referred to the previous studies (Kay, et al. 2008; Naselaris, et al. 2009), and the dataset can be downloaded from http://crcns.org/data-sets/vc/vim-1.

### Problem definition and Bayesian theorem

Inspired by previous work (Naselaris, et al. 2009), we first analyzed the visual reconstruction based on the Bayesian viewpoint. From the perspective of probability model, an encoding model that is used to predict brain response (fMRI voxels) $\mathbf{v}$ from the presented image $\mathbf{s}$, can be represented mathematically by a conditional distribution $P(\mathbf{v}|\mathbf{s})$. By contrary, a visual reconstruction aimed at reconstructing the presented stimuli $\mathbf{s}$ from fMRI voxels $\mathbf{v}$ can be represented by an inverse conditional distribution $P(\mathbf{s}|\mathbf{v})$. In this way, visual reconstruction problem can be solved by calculating the probability that each possible image evoked the measured fMRI voxels, and taking the image with highest probability as the reconstruction. However, the distribution $P(\mathbf{s}|\mathbf{v})$ is difficult to obtain. Under the Bayesian theorem, the posterior distribution $P(\mathbf{s}|\mathbf{v})$ can be changed into $P(\mathbf{s})P(\mathbf{v}|\mathbf{s})$ through $P(\mathbf{s}|\mathbf{v}) \propto P(\mathbf{s})P(\mathbf{v}|\mathbf{s})$, in which Prior probability $P(\mathbf{s})$ requires images natural and conditional distribution $P(\mathbf{r}|\mathbf{s})$ requires images matching with brain response based on the encoding model. In this way, the separation of high naturalness and fidelity for visual reconstruction is achieved, and we can avoid the mutual interfering of fidelity and natural item during retraining the GAN. Thus, we designed a visual reconstruction model based on Bayesian manner to deal with naturalness and fidelity of reconstruction individually.

**Overview of proposed method**

Figure 1 presents the flows of the proposed GAN-BVRM, and the model can be divided into four parts. The generator of BigGAN have the two following inputs: categories and random noise vectors sampled from Gauss noise distribution. The generator can generate various natural images that belong to specified categories according to different sampling of noise distribution. Hence, the category decoding model is required. For part (a), we directly trained a nonlinear classifier from fMRI data to the categories by using bidirectional recurrent neural network (BRNN) from our previous study (Qiao, et al. 2019a). After training, the pre-trained classifier can predict the categories given the corresponding fMRI data. However, the pre-trained GAN is trained based on ImageNet dataset with 1000 categories, and these categories are different from the 10 categories of Gallant data (Naselaris, et al. 2009) used in this study. Thus, we manually mapped the predicted Gallant categories into the 1000 ImageNet categories, and one Gallant category can be mapped to a set of ImageNet categories. Part (b) employed the generator of pre-trained BigGAN to generate natural images according sampled categories from the mapped set of categories and noise vectors sampled from Gauss distribution. From the Figure 1, we can see that the generated natural colour images are different from the pattern of visual stimuli viewed by subjects. Hence, we multiplied the generated images by the grey mask to obtain the Gallant-pattern images. FMRI device acquires the human brain activity (voxels) when subject viewed visual stimuli, and the encoding model is used to simulate the procedure of human visual processing for part (c). Through training an encoding model to construct the mapping from visual stimuli space into fMRI voxel space, we can predict the fMRI voxels from Gallant-pattern images based on the pre-trained encoding model. In this study, we used the pre-trained encoding model for V1, V2, and V3 in our previous work (Qiao, et al. 2019b) because the lower-level visual cortices are mainly responsible for the detailed visual information such as texture, location, and so on. Under the Bayesian framework, we evaluated the similarity between generated images and corresponding visual stimuli by measuring the distance between the true and predicted voxels. According to the above flows, we can evaluate and rank the infinite generated images obtained on the basis of infinite Gauss noise vectors sampled from Gauss distribution. Finally, those images ranked at the front are the reconstruction of the corresponding fMRI data. From the perspective of Bayesian manner, the naturalness of reconstruction can be preserved through the excellent pre-trained generator, semantic content can be determined through decoded categories and the fidelity of reconstruction can be guaranteed through the encoding model. Thus, category decoding is used to control the semantic content and encoding model is used to control the detailed content for visual reconstruction. We employed the PyTorch deep network framework (Ketkar 2017) to perform the experiment based on the Ubuntu 16.04 system with four NVIDIA Titan Xp graphics card. In the experiment, we set batch size as 256 for generator of BigGAN, Approximately 400 iterations (256×400=100K images) are required to accomplish the reconstruction of each visual stimuli.

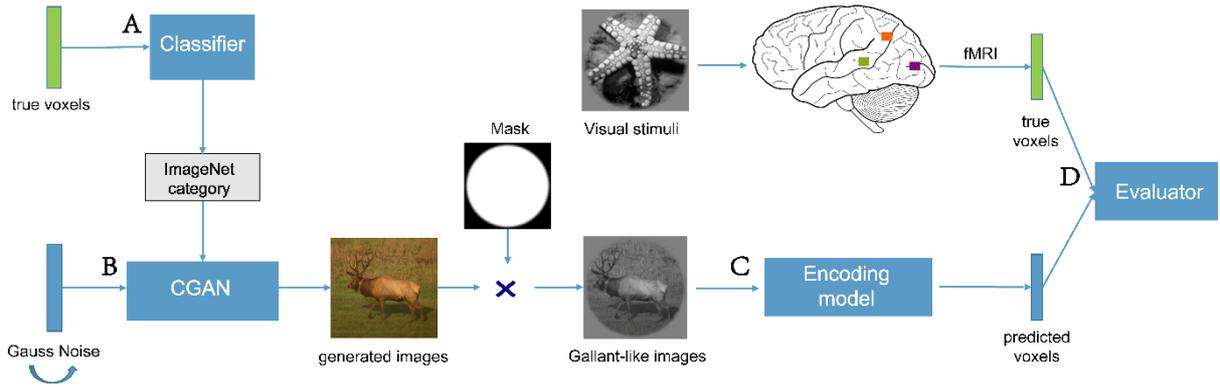

Figure 1. Illustrator of the proposed method. The model mainly includes the four following parts: A. classifier used to decode categories from voxels, B. pre-trained generator of CGAN used to generate natural images, C. encoding model used to encode generated images to predict voxels, and D. evaluator used to evaluate the distance between predicted and true voxels as to measure the quality of generated images.

**Category decoding as prior knowledge of conditional generator**

Common category decoding methods employ a statistical linear or nonlinear classifier to directly learn the mapping from specific voxel patterns in visual cortices to the categories. The internal relationship among different visual areas is ignored, and voxels in selected visual cortices are regarded as a whole to feed into the decoding model. Compared to feedforward neural networks, BRNNs have two directions and can use input information from the past and future of the current time frame for sequence modelling. Thus, as our previous study (Qiao, et al. 2019a), we introduced the bidirectional information flows into our decoding model and constructed an RNN-based model that employs voxels in each visual area as space sequence input to predict categories. The output of the BRNN module can be regarded as the representations of the bottom-up and top-down information flows among visual cortices.

In detail, the input sequence containing five nodes is composed of those voxels selected from five visual areas (V1, V2, V3, V4 and LO), and each node receives the fixed number (100) of effective voxels. After feeding the input sequence to the decoding model, a layer of bidirectional long short term memory (LSTM) module is used to extract the features about categories from the sequence. In this way, it is easier to mine hierarchical features and relationship features between visual areas. The decoding model then combines the output features from two directions and feeds them into the successive fully connected softmax layer to predict categories. In detail, the input node is 100-D, and the output of the node in each direction of LSTM is a 16-D feature. Thus, a 32-D feature combining two directions is obtained for next classification. The number of nodes in the last fully connected softmax layer is 10-D. The dropout operation with a 0.5 rate behind the output of bidirectional LSTM is added to avoid the overfitting. The cross entropy is used to train the decoding model. Further experiment details can refer to the previous study.

After training, the decoding model can predict the category, given the corresponding fMRI voxels from V1, V2, V3, V4, and LO. In the test set, the classification accuracy of decoding model approach 46.17±0.42 for subject 1. However, the image visual stimuli used in this study have 10 categories (many humans, few humans, mammal, non-mammal, non-building, building, plant, non-plant, organic texture, and inorganic texture), which is much too different from the 1000 categories in ImageNet dataset. Given that the condition generator is trained based on ImageNet dataset, we manually constructed the mapping from the Gallant categories (10 categories) to ImageNet categories (1000 categories), and one-to-many correspondence (one Gallant category corresponds to a number of ImageNet categories) is finally obtained. In this way, once the decoded category is obtained, the corresponding ImageNet categories are fed into the next condition generator to generate images.

In detail, ImageNet dataset have different labels from Gallant dataset. ImageNet classes have many different species of fishes (non-mammal animal), birds (non-mammal animal), snakes (non-mammal animal), dogs (mammal animal), and so on, which almost belong to "animal" in Gallant dataset. In addition, ImageNet dataset has enormous artefacts (man-made: non-building) such as refrigerator, bags, displayer, kettle, and so on. By contrast, some Gallant classes such as "humans" and "texture" have fewer corresponding categories in ImageNet dataset, thus, we tried to select some related ImageNet categories for "humans" and "texture" categories. For example, we selected some "sports" category that need many humans to take part in for "many humans" category, some categories that include text texture ("website", "street sign", "menu") for "inorganic texture" category. Even though, these 10 categories still have unbalanced number of corresponds, and we provide the number of classes and some associated with each Gallant category in the Table 1.

Table 1. Corresponds between Gallant and ImageNet categories.

| Index | Gallant categories | ImageNet categories | number |
|---|---|---|---|
| 1 | many humans | ballplayer, hockey puck, soccer ball | 11 |
| 2 | few humans | bridegroom, abaya, bathing cap | 43 |
| 3 | mammal | malinois, briard, tiger, lion | 219 |
| 4 | non-mammal | goldfish, tiger shark, cock, junco | 171 |
| 5 | non-building | ambulance, airship, analog clock | 402 |
| 6 | building | Altar, barn, beacon, bell cote, castle | 54 |
| 7 | plant | broccoli, head cabbage, cauliflower | 41 |
| 8 | non-plant | alp, cliff, lakeshore, sandbar, seashore | 35 |
| 9 | organic texture | honeycomb, velvet, trilobite, brain coral | 12 |
| 10 | inorganic texture | website, street sign, menu, bubble | 12 |

**BigGAN as an infinite image dataset**

BigGAN also has two neural network models and achieves the transferring from noise distribution to image distribution through competing training. The employed BigGAN in this study was trained on ImageNet at 128×128 resolution and achieved the state-of-the-art performance for generating natural images. Based on decoded categories, the generator of pre-trained BigGAN can generate one image that belongs to the specified category based on one sampling from Gauss noise distribution. The specified category can control the semantic content of the generated image, and the values of random noise vector influences the detailed content of the generated image, such as shape, colour, location, and so on. In this way, infinite samplings from Gauss noise distribution can obtain infinite natural images, in which, those images fitting well with fMRI voxels through successive encoding model can be determined as final reconstruction. Hence, the BigGAN plays an important role in the overall method.

BigGAN employed the SA-GAN (Zhang, et al. 2018b) architecture as the baseline and provided category information to the generator with class-conditional batch normalization, and to discriminator with projection. BigGAN simply increased the number of channels in each layer for model architecture and the batch size during training. In detail, the original inputs include one 120-D noise vector and one hot 1000-D category vector. The 120-D noise vector is split into six vectors in equal size, and each 20-D vector is concatenated with the shared category embedding and fed into a corresponding residual block as a conditioning vector. The intuition behind this design is to allow the generator to use the latent space to directly influence the features at different resolutions and levels of hierarchy. Pre-trained generator model of BigGAN used in this study can be downloaded from https://github.com/ajbrock/BigGAN-PyTorch. Further details of BigGAN can refer to original study.

The images viewed by subject were converted into grayscale and masked with circle, whose pattern is different from the generated images. Hence, we similarly converted the generated images into grayscale images and processed them with a circle mask for the next encoding and evaluation.

**Encoding model mapping generated images to voxels**

Infinite images can be generated with infinite noise vectors based on the generator and corresponding categories. Finding suitable images with true voxels at hand is important for the final reconstruction. Generated images and true voxels belong to the two different spaces, and one mapping from images to voxels, namely, one encoding model, is required. Thus, we employed one encoding model to map the generated visual stimuli to corresponding voxels, and then evaluate them by comparing with true voxels. Current encoding models mainly used the two-step manner of encoding including first selecting well matching feature transformation with corresponding brain activity, and second encoding each voxel through linear regression. This type of two-step manner easily falls into the local optimal status and cannot approach global optimal status. In addition, the linear regression model used voxel-wise manner, and one individual linear regression model was trained for each voxel. Eventually, thousands of regression models are constructed for several visual ROIs, which is inefficient. In this study, we employed an effective and efficient encoding model as our previous study (Qiao, et al. 2019b). As the high-level visual areas are difficult to encode and have bad encoding performance, we employed the encoding model for V1, V2, and V3 to preserve the fidelity of visual reconstruction.

In detail, the encoding model was constructed on the basis of one convolution regression model that was trained in an end-to-end manner in which all model parameters are trained jointly, instead of step by step. The encoding model can be divided into two parts, and front convolutional operations (S2F module) was used to extract good matching features with voxels and the last one linear fully connected (fc) layer (F2V module) was used to predict voxels. The S2F module used for feature transformation employed small convolutional kernel (3×3) for V1 and V2, and large convolutional kernel (5×5) for V3. The self-adapting regression weights in fc layer were employed for the voxels to pay more attention to those well-related features with itself. In addition, ROI-wise encoding manner was used to replace the traditional voxel-wise encoding, which also reduces the number of encoding models and makes the reconstruction efficient. Regarding the loss, we employed Person correlation (PC) instead of mean square error (MSE) between the observed and the predicted responses. To avoid the interfering of ineffective voxels during ROI-wise encoding, we used weighted correlation loss and noise regularization to accomplish selective optimization and update all of the parameters of convolution regression model. Further details of the encoding model were described in a previous study (Qiao, et al. 2019b) and corresponding open-source repository (https://github.com/KaiQiao1992/ETECRM).

**Evaluator ranking generated images**

After mapping generated images to voxels space, we evaluated the distance between the predicted and true voxels. However, the encoding model aimed at obtaining high correlation instead of small distance, and the PC value cannot evaluate the single sample. Hence, we trained an additional one-to-one linear mapping with one weight variable and one bias variable to unify the two spaces of the predicted and true voxels. Afterwards, we can employ the MSE to measure the distance between the predicted and true voxels that reflects whether the images accord with the voxels. In this way, we can rank all generated images according to corresponding MSE distances, and smallest MSE value represented the most suitable images. Given that some ineffective voxels influences the computation of MSE, we removed the ineffective voxels and computed the mean values of effective voxels whose encoding performance on training set exceeds 0.27 (Kay, et al. 2008). Essentially, the encoding models for V1, V2, and V3, and the evaluator were used to restrict the detailed content of reconstruction, the decoded category was used to restrict the semantic content. By splitting the reconstruction into two individual parts, GAN-BVRM can avoid the contradiction between fidelity and naturalness and achieved the good fidelity and naturalness of reconstruction.

## Results

**Reconstruction based on random categories**

We first performed visual reconstruction based on the BVRM without decoded categories, and feed random categories into the generator. In this way, the generated images with all categories are equally selected to maximize to fit the encoding model. Given that the encoding models was only encoded for lower-level visual areas, lower-level visual features are only considered in the reconstruction. Figure 2 presents the qualitative results, and the quantitative results (MSE and SSIM) are also given in Table 2. We can see that reconstruction results are consistent with the original visual stimuli regardless of semantic content. For example, the reconstruction of the fifth line of image is similar with the image stimuli in terms of scene that background is shady grove, and foreground is empty land. These results validate the proposed method, and demonstrated the important role of lower-level visual areas in the architecture. In addition, the low-level appearance of the ranked Top10 reconstruction images for each image stimulus shows consistency. This finding indicates the robustness of the BVRM. However, we can see some incoherence, such as the third reconstructions (zebra) of the fifth line of image stimulus, which indicates the lack of the control of right semantic category and the restriction of only lower-level visual areas used for reconstruction. Note that the presented results are from subject 1, and corresponding results for subject 2 can be seen from Appendices, which indicates the similar phenomenon.

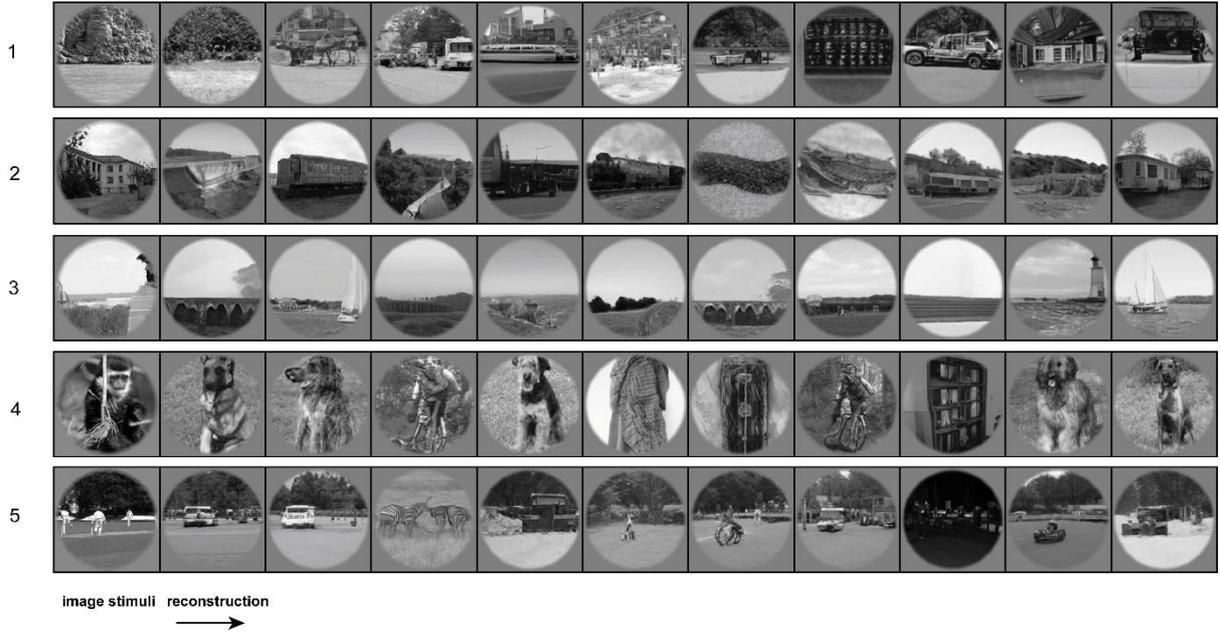

Figure 2. Reconstruction results of five image stimuli. The first column represents the stimuli viewed by subject, and the corresponding ranked Top10 reconstructions are from the second to eleventh column. Note that these ranks are obtained according to distance between predicted voxels and true voxels.

Table 2. Quantitative evaluation based on MSE and SSIM. "Top1" represents the evaluation of ranked first reconstruction that located at the second column, and "Top10" represents the evaluation of ranked Top10 reconstruction that located at from second column to the end.

| Index | 1 | 2 | 3 | 4 | 5 | Mean |
|---|---|---|---|---|---|---|
| MSE (Top1) | 0.035 | 0.033 | 0.037 | 0.034 | 0.068 | 0.089 |
| MSE (Top10) | 0.049 | 0.039 | 0.036 | 0.059 | 0.075 | 0.091 |
| SSIM (Top1) | 0.392 | 0.517 | 0.646 | 0.529 | 0.322 | 0.322 |
| SSIM (Top10) | 0.394 | 0.502 | 0.629 | 0.443 | 0.313 | 0.321 |

**Reconstruction based on predicted categories**

We then presents the reconstruction based on predicted categories in the Figure 3. We can see that the semantic and detailed content are consistent with the original image stimuli. The fidelity and naturalness of reconstruction are both guaranteed. Table 3 gives the corresponding quantitative results. These results indicated that the fine reconstruction can be obtained by combining the semantic categories and low-level visual features. In addition, some false decoded categories exist and influence the semantic content of reconstruction. The proposed method divides the visual reconstruction problem into category decoding and detailed content decoding. For the reconstruction with wrongly predicted categories, we can see the similarity in terms of low-level appearance. With improved decoding category, better reconstruction can be obtained. We also present all reconstruction results for all 120 visual stimuli of testing set in Figure 4 and these results reveals the robustness of the proposed GAN-BVRM.

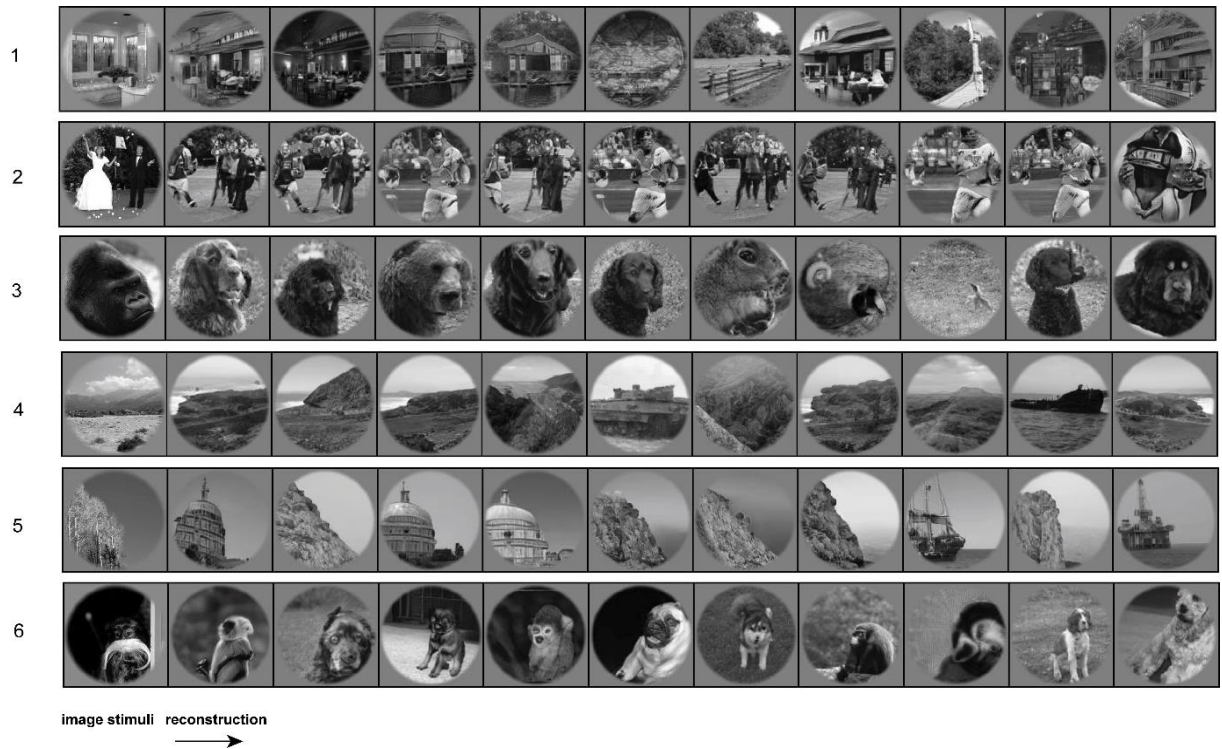

Figure 3. Reconstruction results of six image stimuli. The first column represents the stimuli to be reconstructed from voxels, and the corresponding ranked Top10 reconstructions are from the second column to the end. Note that these ranks are obtained according to distance between predicted and true voxels.

Table 3. Quantitative evaluation based on MSE and SSIM.

| Index | 1 | 2 | 3 | 4 | 5 | 6 | Mean |
|---|---|---|---|---|---|---|---|
| MSE (Top1) | 0.050 | 0.084 | 0.055 | 0.056 | 0.023 | 0.059 | 0.079 |
| MSE (Top10) | 0.062 | 0.099 | 0.059 | 0.052 | 0.045 | 0.091 | 0.078 |
| SSIM (Top1) | 0.355 | 0.339 | 0.369 | 0.479 | 0.614 | 0.298 | 0.343 |
| SSIM (Top10) | 0.333 | 0.318 | 0.396 | 0.488 | 0.545 | 0.286 | 0.342 |

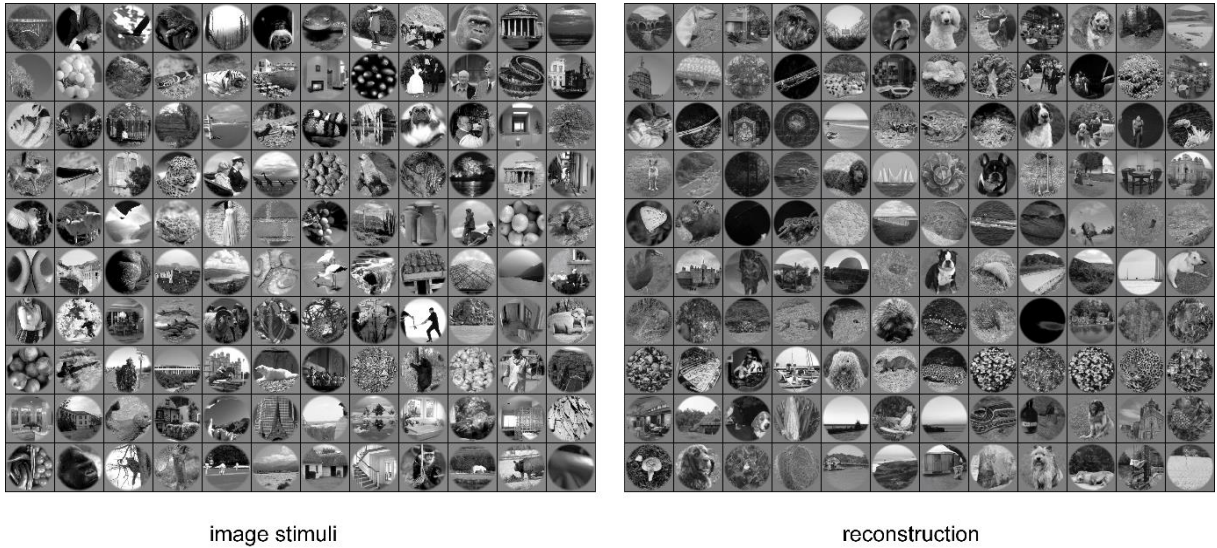

Figure 4. Reconstruction results for 120 image stimuli of testing set.

**Infinite dataset (BigGAN) versus limited dataset (ImageNet dataset)**

For the proposed architecture, BigGAN is an infinite image dataset, and the optimal images that are best matched with the voxels through encoding model were selected as reconstruction images. BigGAN have learned data distribution through ImageNet dataset instead of remembering all images, which have been validated in the original paper. In theory, the BigGAN as an infinite dataset can generate various images and thus is better than the limited ImageNet dataset for visual reconstruction. To prove the effect of BigGAN, we replaced GAN model and used the ImageNet dataset to complete the searching process. Figure 5(a) demonstrates a part of results, and Figure 5(b) presents the quantitative comparison. From these results, we can see that Infinite dataset was better than limited dataset.

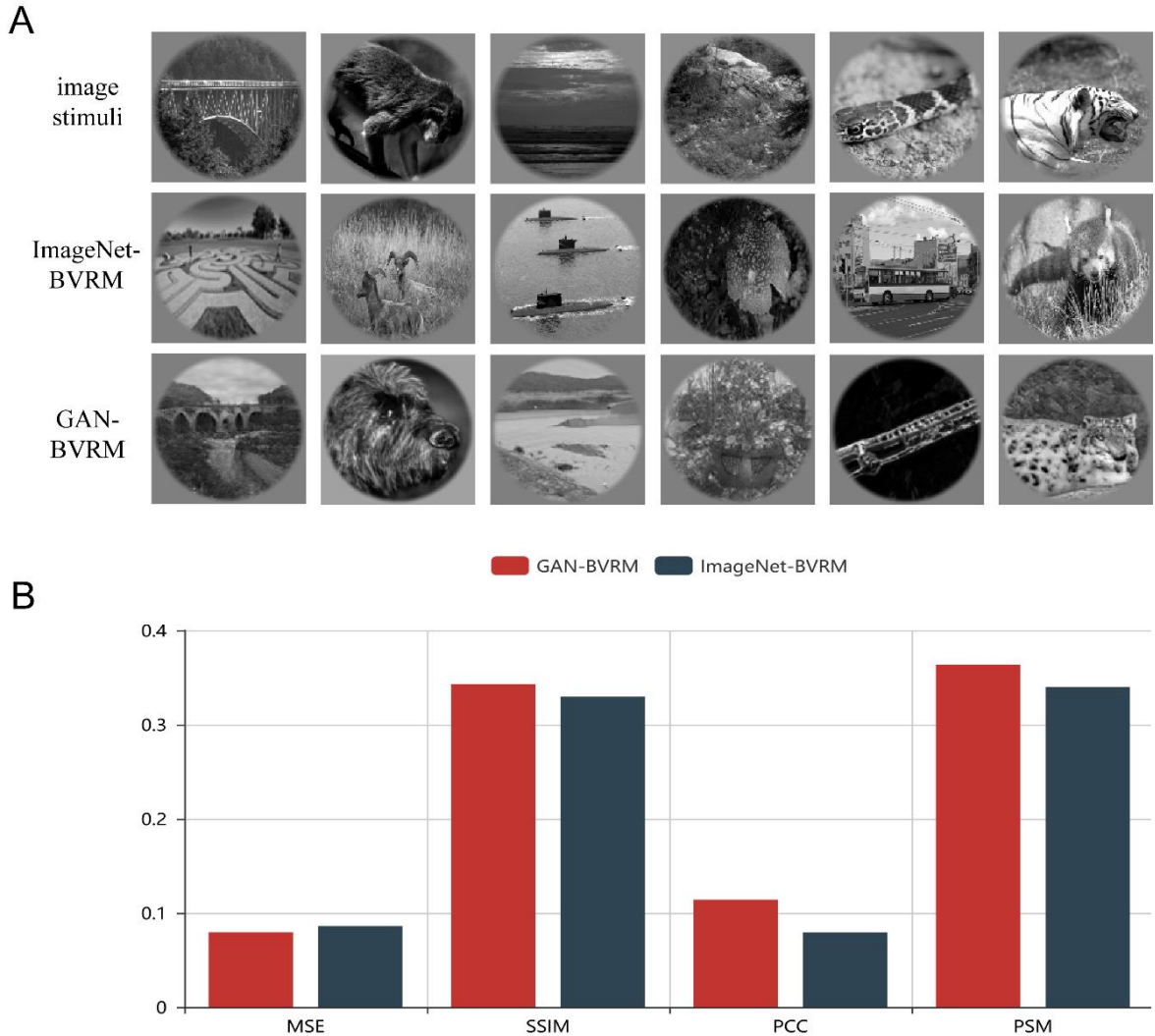

Figure 5. Qualitative and quantitative reconstruction results based on BigGAN and ImageNet dataset.

**Comparison with current GAN-based method**

We also provided comparing results with the two recently GAN-based method (Seeliger, et al. 2018; St-Yves and Naselaris 2018), and the two methods are called for short GAN-M1 and GAN-M2. All results for the two method in the Figure 5 are obtained from original papers. From the Figure, we can see that the GAN-BVRM obtain better reconstruction than the two methods. In detail, the reconstruction images of the first method are 64×64, and the second method only reconstruct 32×32 image. These images (32×32) are difficult to determine regardless of fidelity and naturalness. The reconstructed images from the first method have slight fidelity, but reconstruction results are difficult to identify because of worse naturalness. By contrast, the GAN-BVRM performs well in terms of fidelity and naturalness, thus confirming its effectiveness. GAN-BVRM can generate 128×128 images, and the scale was difficult for current methods, including GAN-based methods, because generating big sized images was difficult for GAN training with minority of data.

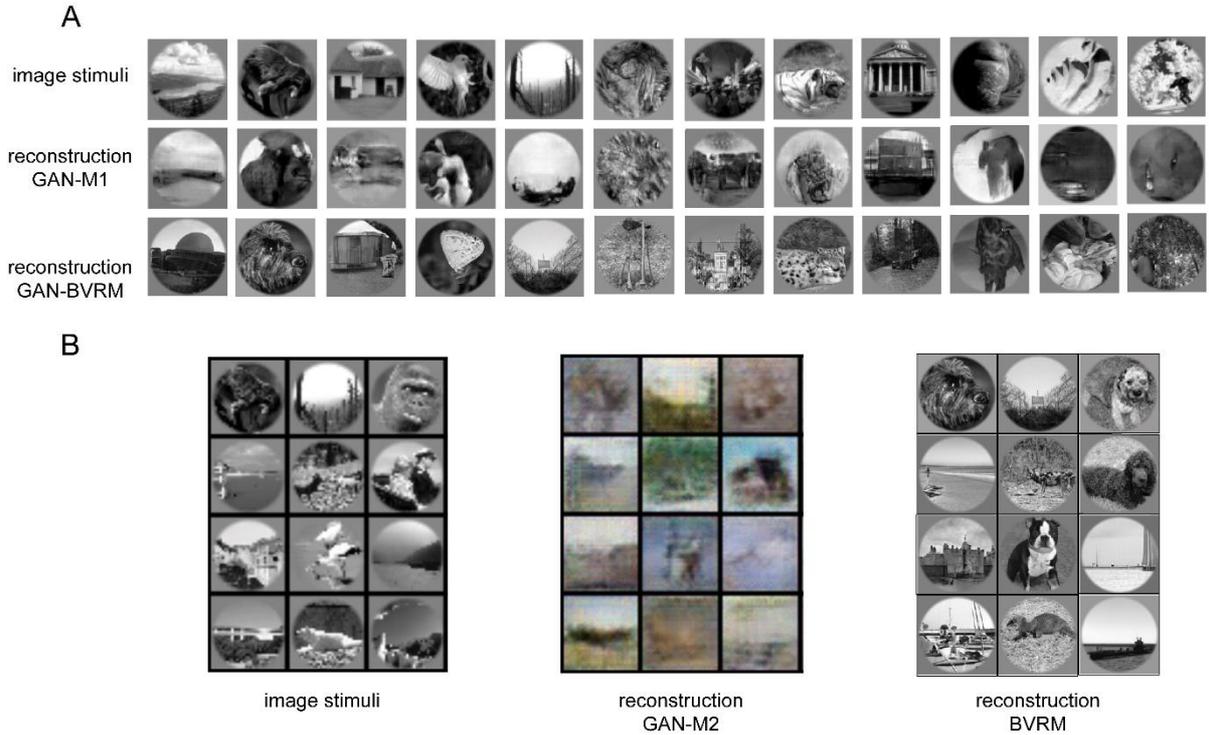

Figure 6. Comparison of the two GAN-based methods, namely, "GAN-M1" (Seeliger, et al. 2018) in A and "GAN-M2" (St-Yves and Naselaris 2018) in B.

In addition, we used average MSE, structure similarity index metric (SSIM), Pearson correlation coefficient (PCC) and perceptual similarity metrics (PSM) to quantitatively compare the reconstruction results of our method and GAN-M1. These results of GAN-M1 are generously provided by Seeliger, et al (Seeliger, et al. 2018). As shown in the Figure 7, we can see that our method perform better performance according to the four metrics. For one image, its different layers' CNN features represents different level of appearance. PSM is recently proposed to evaluate the similarity on these different level of appearance. The similarity on the high-level features and low-level features represents the similarity of detailed and semantic appearance, respectively. Thus, we employed AlexNet model (Krizhevsky, et al. 2012) to obtain each layers' features of reconstruction images and presented images, and computed the correlation for each layer. In the Figure 8, we provided the detailed similarity from fist Conv layer to last Conv layer including middle ReLU layer and Polling layer. The results demonstrated that our reconstruction results were more similar with original images from low-level to high-level features. Thus, these qualitative and quantitative reconstruction results indicated that GAN-BVRM with high robustness improved the visual reconstruction quality.

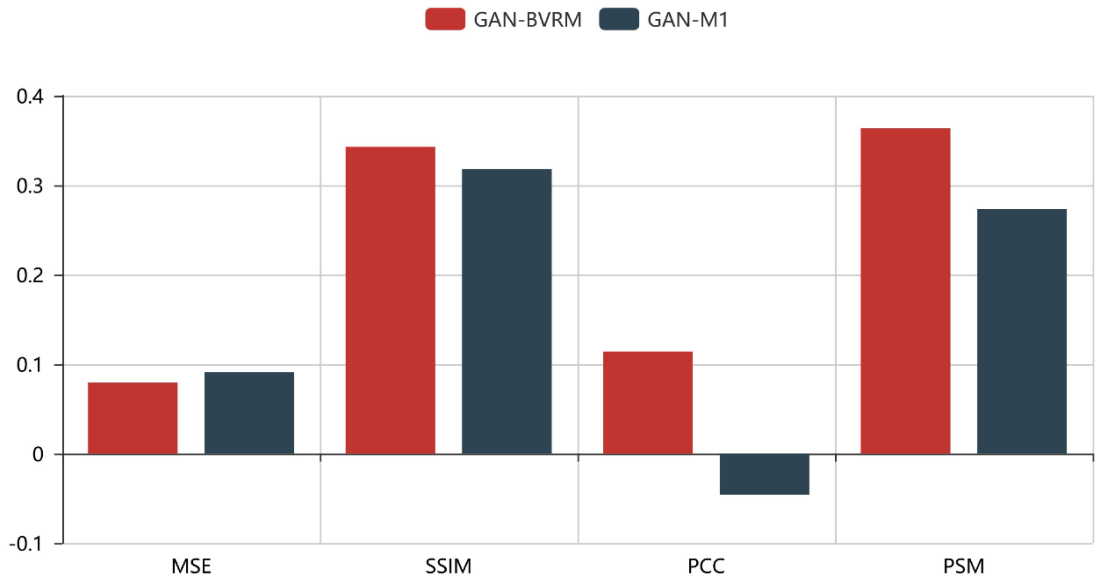

Figure 7. Comparison between GAN-BVRM and GAN-M1 based on several metrics.

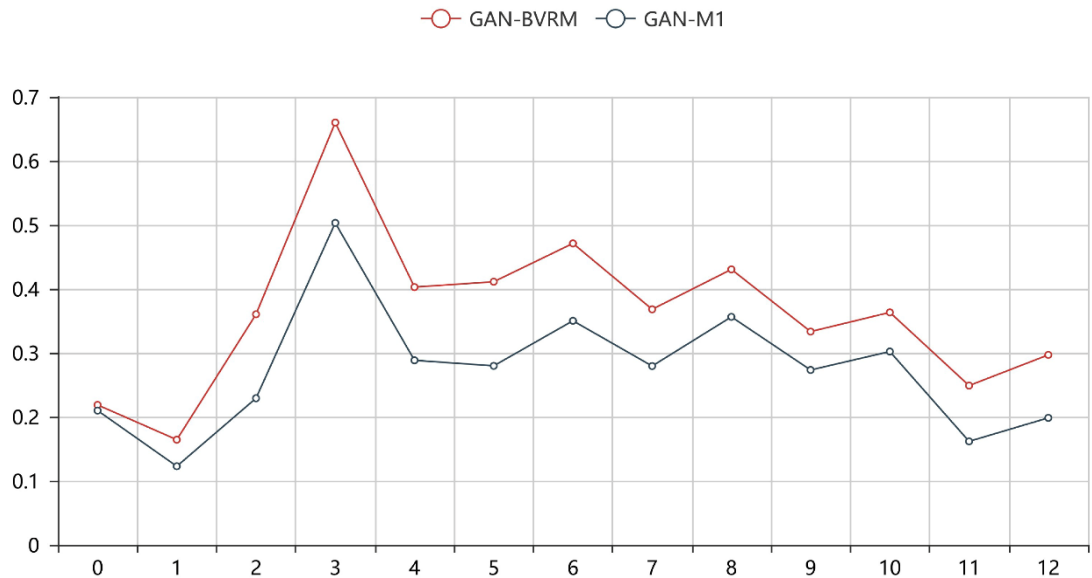

Figure 8. Perceptual performance and comparison based on different level of CNN features.

## Discussion

**Advantage and disadvantage**

In this study, we proposed the GAN-BVRM by maximizing to fit with the encoding model using BigGAN in the Bayesian framework. Through visual encoding and category decoding module, we accomplished the reconstruction with the infinite image database constructed by the pre-trained generator of BigGAN. In this way, our proposed method provides a new architecture for visual reconstruction based on GAN. The proposed method successfully splits the reconstruction problem into several subproblems to avoid the contradiction of fidelity and naturalness from training one GAN-based model from starch. The pre-trained generator preserves the naturalness, and the encoding models of lower-level visual areas preserve the fidelity, in which the two parts are not cross. In addition, the size of our reconstruction images reach 128×128 which is an acceptable scale, compared to previous 64×64 or even 32×32 results. Certainly, several parts still need to be improved, for example, the searching-based manner has low efficiency, and non-uniform categories between Gallant and ImageNet dataset still constricts the performance of GAN-BVRM. Given that all of the four parts are fully constructed based on neural networks, the proposed method is potential to achieve the back propagation, and iteratively update the noise vector to generate optimal images to efficiently accomplish visual reconstruction.

**Diversity of image generating**

In the proposed reconstruction paradigm, the pre-trained generator of BigGAN is regarded as infinite image database and plays an important role in the reconstruction. However, BigGAN has exposed that generated images have weak diversity (Razavi, et al. 2019) for many reasons, such as the truncation trick that truncates the large values of noise vector to improve the quality but sacrifice the variety. In this way, the infinite image dataset has the problem of diversity, some optimal images for the encoding model may be missed, and the reconstruction may fall into the local optimal in the constricted dataset. On the other hand, it implies the potential of the proposed method. Hence, improved GAN can directly advance the performance of reconstruction through the proposed GAN-based Bayesian framework, which is also benefit from the proposed architecture.

**Replacement of the encoding model**

In the proposed method, encoding model is used to map visual stimuli to voxels, namely, simulating human visual processing. In this way, we can present the generated images to the subject, and monitor brain activity with fMRI and then compare it with corresponding previous voxels of visual stimulus to be decoded, finally we can evaluate the generated image. Human vision system can replace the encoding model and certainly have better encoding performance. The manner is similar with recently proposed method that uses GAN to generate those images to activate the specific neural voxel (Bashivan, et al. 2019). Through the manner, the proposed method can interact with the subject to accomplish better reconstruction.

# Conclusion

Current GAN-based methods for visual reconstruction cannot approach the fine fidelity and naturalness. In this study, we proposed the GAN-BVRM by using pre-trained image generator, category decoding, encoding mapping and evaluator in the Bayesian framework. By maximizing to fit the encoding model, the most suitable images from the infinite database constructed by the generator of pre-trained BigGAN, can be selected as the reconstruction. Experimental results demonstrated that the GAN-BVRM preserved the fidelity and naturalness. We also analyzed the GAN-BVRM in terms of diversity of generator, improved encoding performance, and category decoding, which demonstrated the potential for visual reconstruction based pre-trained GAN in the Bayesian manner.

## Data Availability

The detailed information about the fMRI data is described in previous studies (Kay, et al. 2008; Naselaris, et al. 2009), and the public dataset can be downloaded from *http://crcns.org/data-sets/vc/vim-1*.

## Conflicts of Interest

The authors declare that there is no conflict of interest regarding the publication of this paper.

## Funding Statement

This work was supported by the National Key Research and Development Plan of China (No. 2017YFB1002502), the National Natural Science Foundation of China (No. 61701089), and the Natural Science Foundation of Henan Province of China (No. 162300410333).

## Appendices A. Results for subject 2

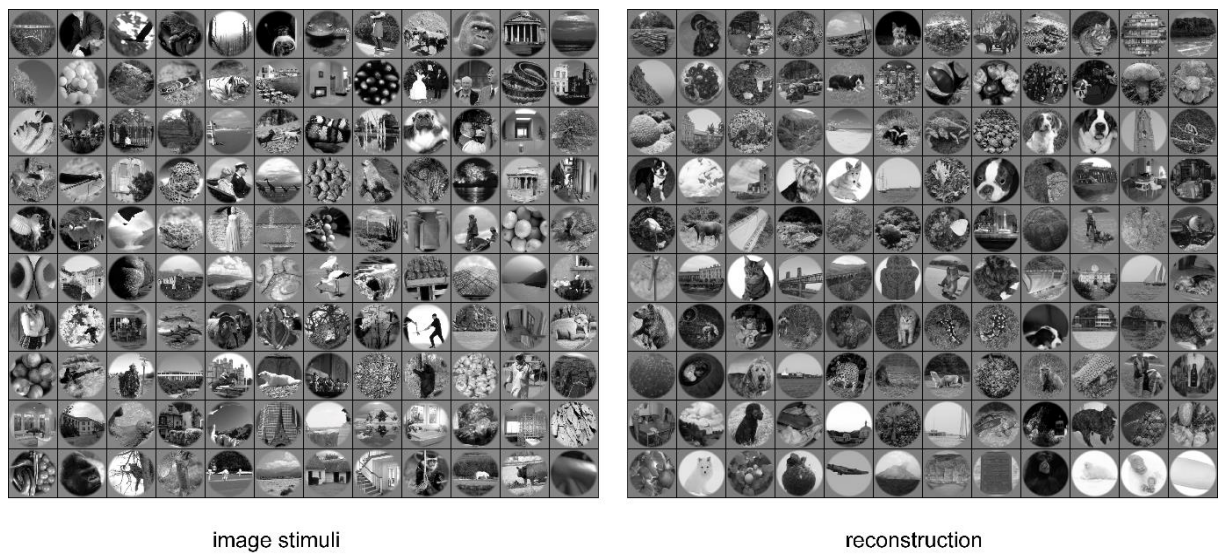

image stimuli                                              reconstruction

Figure 9. Reconstruction results for 120 image stimuli of testing set for the subject 2.

## References

Arjovsky, M., Chintala, S., Bottou, L. (2017) Wasserstein generative adversarial networks. International Conference on Machine Learning. p 214-223.
Bashivan, P., Kar, K., DiCarlo, J.J. (2019) Neural population control via deep image synthesis. Science, 364:eaav9436.
Berthelot, D., Schumm, T., Metz, L. (2017) BEGAN: boundary equilibrium generative adversarial networks. arXiv preprint arXiv:1703.10717.
Brock, A., Donahue, J., Simonyan, K. (2018) Large scale gan training for high fidelity natural image synthesis. arXiv preprint arXiv:1809.11096.
Chen, X., Duan, Y., Houthooft, R., Schulman, J., Sutskever, I., Abbeel, P. (2016) Infogan: Interpretable representation learning by information maximizing generative adversarial nets. Advances in neural information processing systems. p 2172-2180.

Cox, D.D., Savoy, R.L. (2003) Functional magnetic resonance imaging (fMRI)"brain reading": detecting and classifying distributed patterns of fMRI activity in human visual cortex. Neuroimage, 19:261-270.
Eickenberg, M., Gramfort, A., Varoquaux, G., Thirion, B. (2016) Seeing it all: Convolutional network layers map the function of the human visual system. Neuroimage, 152.
Goodfellow, I., Pouget-Abadie, J., Mirza, M., Xu, B., Warde-Farley, D., Ozair, S., Courville, A., Bengio, Y. (2014) Generative adversarial nets. Advances in neural information processing systems. p 2672-2680.
Güçlü, U., van Gerven, M.A. (2015) Deep neural networks reveal a gradient in the complexity of neural representations across the ventral stream. Journal of Neuroscience, 35:10005-10014.
Han, K., Wen, H., Shi, J., Lu, K.-H., Zhang, Y., Liu, Z. (2017) Variational autoencoder: An unsupervised model for modeling and decoding fMRI activity in visual cortex. bioRxiv:214247.
Haxby, J.V., Gobbini, M.I., Furey, M.L., Ishai, A., Schouten, J.L., Pietrini, P. (2001) Distributed and overlapping representations of faces and objects in ventral temporal cortex. Science, 293:2425-2430.
Haynes, J.-D., Rees, G. (2006) Neuroimaging: decoding mental states from brain activity in humans. Nature Reviews Neuroscience, 7:523.
Horikawa, T., Kamitani, Y. (2017) Hierarchical neural representation of dreamed objects revealed by brain decoding with deep neural network features. Frontiers in computational neuroscience, 11:4.
Horikawa, T., Tamaki, M., Miyawaki, Y., Kamitani, Y. (2013) Neural decoding of visual imagery during sleep. Science, 340:639-642.
Hossein-Zadeh, G.-A. (2016) Reconstruction of digit images from human brain fMRI activity through connectivity informed Bayesian networks. Journal of neuroscience methods, 257:159-167.
Huth, A.G., Nishimoto, S., Vu, A.T., Gallant, J.L. (2012) A continuous semantic space describes the representation of thousands of object and action categories across the human brain. Neuron, 76:1210-1224.
Jiao, Z., You, H., Yang, F., Li, X., Zhang, H., Shen, D. (2019) Decoding EEG by visual-guided deep neural networks. Proceedings of the 28th International Joint Conference on Artificial Intelligence: AAAI Press. p 1387-1393.
Kamitani, Y., Tong, F. (2005) Decoding the visual and subjective contents of the human brain. Nature neuroscience, 8:679.
Kay, K.N., Naselaris, T., Prenger, R.J., Gallant, J.L. (2008) Identifying natural images from human brain activity. Nature, 452:352.
Ketkar, N. (2017) Introduction to pytorch. Deep Learning with Python: Springer. p 195-208.
Krizhevsky, A., Sutskever, I., Hinton, G.E. (2012) ImageNet classification with deep convolutional neural networks. International Conference on Neural Information Processing Systems. p 1097-1105.
Lin, Y., Li, J., Wang, H. (2019) DCNN-GAN: Reconstructing Realistic Image from fMRI. arXiv preprint arXiv:1901.07368.
Logothetis, N.K., Sheinberg, D.L. (1996) Visual object recognition. Annual review of neuroscience, 19:577-621.
Mahendran, A., Vedaldi, A. (2015) Understanding deep image representations by inverting them. Proceedings of the IEEE conference on computer vision and pattern recognition. p 5188-5196.
Mirza, M., Osindero, S. (2014) Conditional generative adversarial nets. arXiv preprint arXiv:1411.1784.
Mitchell, T.M., Hutchinson, R., Niculescu, R.S., Pereira, F., Wang, X., Just, M., Newman, S. (2004) Learning to decode cognitive states from brain images. Machine learning, 57:145-175.
Mitchell, T.M., Shinkareva, S.V., Carlson, A., Chang, K.-M., Malave, V.L., Mason, R.A., Just, M.A. (2008) Predicting human brain activity associated with the meanings of nouns. science, 320:1191-1195.
Miyawaki, Y., Uchida, H., Yamashita, O., Sato, M.-a., Morito, Y., Tanabe, H.C., Sadato, N., Kamitani, Y. (2008) Visual image reconstruction from human brain activity using a combination of multiscale local image decoders. Neuron, 60:915-929.
Naselaris, T., Kay, K.N., Nishimoto, S., Gallant, J.L. (2011) Encoding and decoding in fMRI. Neuroimage, 56:400-10.
Naselaris, T., Prenger, R.J., Kay, K.N., Oliver, M., Gallant, J.L. (2009) Bayesian Reconstruction of Natural Images from Human Brain Activity: Neuron. Neuron, 63:902-915.
Norman, K.A., Polyn, S.M., Detre, G.J., Haxby, J.V. (2006) Beyond mind-reading: multi-voxel pattern analysis of fMRI data. Trends in cognitive sciences, 10:424-430.
Ogawa, S., Lee, T.-M., Kay, A.R., Tank, D.W. (1990) Brain magnetic resonance imaging with contrast dependent on blood oxygenation. Proceedings of the National Academy of Sciences, 87:9868-9872.
Palazzo, S., Spampinato, C., Kavasidis, I., Giordano, D., Shah, M. (2017) Generative adversarial networks conditioned by brain signals. Proceedings of the IEEE International Conference on Computer Vision. p 3410-3418.
Qiao, K., Chen, J., Wang, L., Zhang, C., Zeng, L., Tong, L., Yan, B. (2019a) Category decoding of visual stimuli from human brain activity using a bidirectional recurrent neural network to simulate bidirectional information flows in human visual cortices. arXiv preprint arXiv:1903.07783.

Qiao, K., Zhang, C., Chen, J., Wang, L., Tong, L., Yan, B. (2019b) Effective and efficient ROI-wise visual encoding using an end-to-end CNN regression model and selective optimization. arXiv e-prints.

Qiao, K., Zhang, C., Wang, L., Chen, J., Zeng, L., Tong, L., Yan, B. (2018) Accurate reconstruction of image stimuli from human functional magnetic resonance imaging based on the decoding model with capsule network architecture. Frontiers in neuroinformatics, 12.

Razavi, A., Oord, A.v.d., Vinyals, O. (2019) Generating Diverse High-Fidelity Images with VQ-VAE-2. arXiv preprint arXiv:1906.00446.

Schmidhuber, J. (2015) Deep learning in neural networks: An overview. Neural networks, 61:85-117.

Schoenmakers, S., Barth, M., Heskes, T., van Gerven, M. (2013) Linear reconstruction of perceived images from human brain activity. NeuroImage, 83:951-961.

Seeliger, K., Güçlü, U., Ambrogioni, L., Güçlütürk, Y., van Gerven, M. (2018) Generative adversarial networks for reconstructing natural images from brain activity. NeuroImage, 181:775-785.

Shen, G., Dwivedi, K., Majima, K., Horikawa, T., Kamitani, Y. (2018) End-to-end deep image reconstruction from human brain activity. bioRxiv:272518.

St-Yves, G., Naselaris, T. (2018) Generative Adversarial Networks Conditioned on Brain Activity Reconstruct Seen Images. bioRxiv:304774.

Thirion, B., Duchesnay, E., Hubbard, E., Dubois, J., Poline, J.-B., Lebihan, D., Dehaene, S. (2006) Inverse retinotopy: inferring the visual content of images from brain activation patterns. Neuroimage, 33:1104-1116.

Tirupattur, P., Rawat, Y.S., Spampinato, C., Shah, M. (2018) ThoughtViz: Visualizing human thoughts using generative adversarial network. 2018 ACM Multimedia Conference on Multimedia Conference: ACM. p 950-958.

Ungerleider, L.G. (1982) Two cortical visual systems. Analysis of visual behavior:549-586.

van Gerven, M.A., de Lange, F.P., Heskes, T. (2010) Neural decoding with hierarchical generative models. Neural computation, 22:3127-3142.

Wen, H., Shi, J., Chen, W., Liu, Z. (2018) Deep Residual Network Predicts Cortical Representation and Organization of Visual Features for Rapid Categorization. Scientific reports, 8:3752.

Wen, H., Shi, J., Zhang, Y., Lu, K.-H., Cao, J., Liu, Z. (2017) Neural encoding and decoding with deep learning for dynamic natural vision. Cerebral Cortex:1-25.

Yamins, D.L., Hong, H., Cadieu, C.F., Solomon, E.A., Seibert, D., DiCarlo, J.J. (2014) Performance-optimized hierarchical models predict neural responses in higher visual cortex. Proceedings of the National Academy of Sciences, 111:8619-8624.

Zeiler, M.D., Taylor, G.W., Fergus, R. (2011) Adaptive deconvolutional networks for mid and high level feature learning. ICCV. p 6.

Zhang, C., Qiao, K., Wang, L., Tong, L., Hu, G., Zhang, R.-Y., Yan, B. (2019) A visual encoding model based on deep neural networks and transfer learning for brain activity measured by functional magnetic resonance imaging. Journal of Neuroscience Methods:108318.

Zhang, C., Qiao, K., Wang, L., Tong, L., Zeng, Y., Yan, B. (2018a) Constraint-free natural image reconstruction from fMRI signals based on convolutional neural network. Frontiers in human neuroscience, 12.

Zhang, H., Goodfellow, I., Metaxas, D., Odena, A. (2018b) Self-Attention Generative Adversarial Networks. arXiv preprint arXiv:1805.08318.

Zhao, J., Mathieu, M., LeCun, Y. (2016) Energy-based generative adversarial network. arXiv preprint arXiv:1609.03126.

Zhu, J.-Y., Park, T., Isola, P., Efros, A.A. (2017) Unpaired image-to-image translation using cycle-consistent adversarial networks. Proceedings of the IEEE international conference on computer vision. p 2223-2232.